# A Novel Online Real-time Classifier for Multi-label Data Streams

Rajasekar Venkatesan[1]*,
Meng Joo Er[1,2]
[1]School of EEE
Nanyang Technological University
Singapore
RAJA0046@e.ntu.edu.sg
EMJER@ntu.edu.sg

Shiqian Wu
[2]School of Machinery and Automation
Wuhan University of Science and Technology
China
shiqian.wu@wust.edu.cn

Mahardhika Pratama
[3]Department of Computer Science and IT
La Trobe University
Australia
m.pratama@latrobe.edu.au

*Abstract*—In this paper, a novel extreme learning machine based online multi-label classifier for real-time data streams is proposed. Multi-label classification is one of the actively researched machine learning paradigm that has gained much attention in the recent years due to its rapidly increasing real world applications. In contrast to traditional binary and multi-class classification, multi-label classification involves association of each of the input samples with a set of target labels simultaneously. There are no real-time online neural network based multi-label classifier available in the literature. In this paper, we exploit the inherent nature of high speed exhibited by the extreme learning machines to develop a novel online real-time classifier for multi-label data streams. The developed classifier is experimented with datasets from different application domains for consistency, performance and speed. The experimental studies show that the proposed method outperforms the existing state-of-the-art techniques in terms of speed and accuracy and can classify multi-label data streams in real-time.

## I. INTRODUCTION

Classification in machine learning is the problem of identifying the function f(x) that maps each attribute vector $x_i$ to its associated target label $y_i$, i = 1,2,….,n, where n is the total number of training samples [1]. Traditional classification problems in machine learning involve associating each of the sample instance with a single target label. i.e. unique target association. This type of classification is called single label classification. On the contrary, several real world classification problems involve data samples which correspond to a subset of target labels. This results in the emergence of a new category of machine learning classification called the multi-label classification. The multi-label classification problems are gaining much importance and attention in the recent years due to the rapidly increasing real world application areas. Some of the real world application domains that require multi-label classification are medical diagnosis [2, 3], text categorization [4-8], genomics, bioinformatics [9, 10], multimedia, emotion, music categorization, scene and video categorization [11-13], map labeling [14], marketing etc. Due to the omnipresence of multi-label problems in a wide range of real world scenarios, multi-label classification is an emerging field in machine learning classification [15].

The traditional single label classification problems maps each of the input samples to a unique target label from the pool of available target labels. The single label classification problems can be categorized into binary and multi-class classification. When the number of available target labels is two, it is called binary classification. Binary classification is the most fundamental classification problem in which the input sample belongs to either of the two target class labels. Examples of binary classification problems include biometric security, medical diagnosis, etc. When the number of available target labels is greater than two, the classification problem is called multi-class classification. Biometric identification, character recognition and other similar classification problems are examples of multi-class classification. Binary classification is a special case of multi-class classification in which the number of target labels is two.

There are several real world applications in which the target labels are not mutually exclusive and requires the need for multi-label classification. Multi-label classification involves associating each of the input samples with a set of target labels. Therefore, multi-label classification forms the superset of binary and multi-class classification problems. When compared to single label classification, multi-label classification is more difficult and more complex due to the increased generality of the multi-label problems [16].

Several machine learning techniques is available in the literature for multi-label classification problems. The existing multi-label classifiers available in the literature are based on Support Vector Machines (SVM), Decision Trees (DT), Extreme Learning Machines (ELM) etc. The machine learning techniques available can be broadly categorized into two categories: Batch learning and Online learning. Batch learning techniques involve collection of all the data samples in prior and estimating the system parameters by processing all the data concurrently. Batch learning techniques require all the training data beforehand and cannot learn from streaming data. This poses a major limitation to the applications of batch learning techniques as several real world applications require learning from sequentially streaming data samples. Online learning is a family of machine learning techniques in which the learning is achieved by incrementally updating the system parameters from the data that arrives sequentially using single-pass learning procedure [17, 18]. Therefore, online learning techniques are preferred over batch learning techniques for real world applications [19, 20].

It is to be noted that there is very limited research on multi-label classification for streaming data applications [21]. Online techniques for multi-label classification are much to be explored. An ELM based online multi-label classifier is proposed for streaming data applications. The proposed ELM based online multi-label classifier outperforms the existing classifiers in speed and performance and also scalable for large scale streaming data applications.

The organization of the rest of the paper is as follows. Section 2 describes the preliminary discussion on multi-label classifiers and extreme learning machines. The details of the proposed method are elaborated in Section 3. The benchmark evaluation metrics and the experimentation specifications used for analyzing the proposed method are described in Section 4. Section 5 summarizes the experimentation results and performance comparison with state-of-the-art techniques and concluding remarks are given in Section 6.

## II. BACKGROUND AND PRELIMINARIES

### A. Multi-label Classifiers

The summary of multi-label learning problem is given as,

— The input space X is of feature dimension D

$x_i \in X$, $x_i = (x_{i1}, x_{i2}, \ldots x_{iD})$

— The label space L is of dimension M

$L = \{\zeta_1, \zeta_2, \ldots, \zeta_M\}$

— Each of the N training samples can be represented by a pair of tuples (input space and label space)

$\{(x_i, y_i) \mid x_i \in X, y_i \in Y, Y \subseteq L, 1 \leq i \leq N\}$

— A training model that maps the input tuple to output tuple.

Sorower [22] defines multi-label classification as, "Given a training set, $S = (x_i, y_i)$, $1 \leq i \leq n$, consisting of n training instances, $(x_i \in X, y_i \in Y)$ drawn from an unknown distribution D, the goal of multi-label learning is to produce a multi-label classifier $h:X \rightarrow Y$ that optimizes some specific evaluation function or loss function".

If there are M target class labels, and $p_i$ denotes the probability that the input sample is assigned to $i^{th}$ class, as opposed to the single label classification in which each of the input samples belongs to only one target label and the set of target labels are mutually exclusive, multi-label classification enables association of multiple labels to the input sample. Therefore, the following inequality holds true for multi-label classification.

$$\sum_{i=1}^{M} p_i \geq 1 \qquad (1)$$

The existing multi-label classification techniques available in literature can be classified into two major categories: Batch learning techniques and online learning techniques. In literature, only a very limited number of application specific online multi-label techniques are available. An overview of exiting techniques is shown in Fig. 1.

### 1) Batch Learning Methods

The multi-label classification techniques available in the literature are largely batch learning based methods. Tsoumakas et. al [23] categorized the existing batch learning based multi-label classification algorithm available in literature into two categories: Problem Transformation (PT) methods and Algorithm Adaptation (AA) methods. Madjarov et. al [24] extended the classification to include Ensemble (EN) based multi-label classification methods.

**PT methods** transform the multi-label classification problems into multiple single-label classification problems. Existing single label classifiers are then used to perform the classification and finally the results from the multiple single label classifiers are combined together to provide the multi-label classification result. There are three sub-categories to the PT methods: Binary relevance methods, Pairwise methods and Label powerset method. **AA methods** extend the base classification algorithm to adapt multi-label problems. AA methods are algorithm-dependent methods. **EN methods** use an ensemble of PT and AA methods to achieve multi-label classification.

### 2) Online Learning Methods

There are limited number of techniques available in the literature on multi-label classification for data streams [21]. A simpler approach is to use batch learning classifiers that trains on new batches of data streams by replacing the classifiers of previous batches. This type of learning is called batch-incremental learning. The first work on multi-label classifier for data streams is based on ensemble of classifiers which are trained on successive data chunks [25]. The paper by Read et. al [26] proposes multi-label stream classification by extending the heoffding tree [27] by using batch multi-label classifier in each node. Spyromitros-Xioufis [28] proposes binary relevance and kNN based multi-label classifier for data streams. Microsoft [29] developed an Active Learning framework for multi-label classification as the result of the increase in demand for the need of multi-label classification in real world multi-media datasets. A Passive-Aggressive method is proposed by Crammer et. al [30] for multi-label classification. A Bayesian Online Multi-label Classification (BOMC) method is developed by Zhang et. al [31] for online multi-label classification. The Passive Aggressive and the Bayesian Online Multi-label Classification techniques are application specific and are implemented only for text categorization datasets.

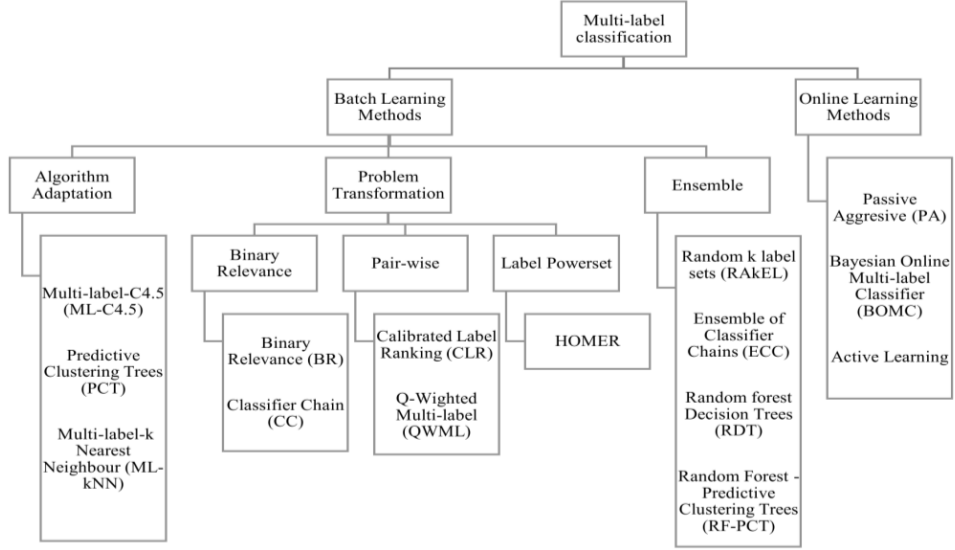

Fig. 1. Overview of Multi-label Classification Techniques

*B. Extreme Learning Machines*

ELM is a single-hidden layer feedforward neural network based learning technique. The special feature of ELM is that the initial weights and the hidden layer bias can be selected at random. This results in high speed training and small number of tunable parameters thus enabling ELM to have fast learning speed and generalization of performance. The universal approximation capability and generalization ability [32] are the key distinguishing factors of ELM. Several variants of ELM has been developed [33-36]. A condensed overview of ELM algorithm as adapted from [32, 37] is discussed.

Consider there are N training samples in the multi-label dataset. Let L be the target label space represented as L = {$\zeta_1$, $\zeta_2$,...., $\zeta_M$}. Consider the input is of the form, $x_i$ = $[x_{i1}, x_{i2}, ..., x_{in}]^T \in R^n$ and the corresponding output is represented as $y_i = [y_{i1}, y_{i2}, ... y_{im}]^T$ and Y⊆L. Let $\tilde{N}$ be the number of hidden layer neurons, the output 'o' of the single hidden layer feedforward neural network is given by

$$\sum_{i=1}^{\tilde{N}} \beta_i g_i(x_j) = \sum_{i=1}^{\tilde{N}} \beta_i g(w_i \cdot x_j + b_i) = o_j \qquad (2)$$

where, $w_i = [w_{i1}, w_{i2}, ... w_{in}]^T$ is the input weight, g(x) is the activation function, and $b_i$ is the hidden layer bias and $\beta_i = [\beta_{i1}, \beta_{i2}, ... \beta_{im}]^T$ is the output weight.

The network should be trained such that the error difference between the actual output and the predicted output is 0.

$$\sum_{j=1}^{\tilde{N}} \|o_j - y_j\| = 0 \qquad (3)$$

Thus, the output of the ELM classifier is given as,

$$\sum_{i=1}^{\tilde{N}} \beta_i g(w_i \cdot x_j + b_i) = y_j \qquad (4)$$

The equation (4) written in matrix form is represented as,

$$H\beta = Y \qquad (5)$$

where,

$$H = \begin{bmatrix} g(w_1 \cdot x_1 + b_1) & \cdots & g(w_{\tilde{N}} \cdot x_1 + b_{\tilde{N}}) \\ \vdots & \ddots & \vdots \\ g(w_1 \cdot x_N + b_1) & \cdots & g(w_{\tilde{N}} \cdot x_N + b_{\tilde{N}}) \end{bmatrix}_{N \times \tilde{N}} \qquad (6)$$

$$\beta = \begin{bmatrix} \beta_1^T \\ \vdots \\ \beta_{\tilde{N}}^T \end{bmatrix}_{\tilde{N} \times m} \qquad (7)$$

$$Y = \begin{bmatrix} y_1^T \\ \vdots \\ y_N^T \end{bmatrix}_{N \times m} \qquad (8)$$

The output weights of the ELM network can be estimated using the equation

$$\beta = H^+ Y \qquad (9)$$

where $H^+$ is the Moore-Penrose inverse of the hidden layer output matrix H, and it can be calculated as follows:

$$H^+ = (H^T H)^{-1} H^T \qquad (10)$$

There are several papers [32, 37, 38] available in literature that elaborates on the theory and the mathematical background behind the ELM and hence are not discussed here. There are other similar neural network based techniques [39, 40] which did not gain popularity and are largely forgotten.

## III. PROPOSED APPROACH

This paper exploits the inherent high speed nature of the ELM and OS-ELM to develop an online sequential multi-label classifier for real-time streaming data applications. The key novelty of the proposed approach is that, there are no online techniques available thus far in literature to perform real-time multi-label classification.

In single label classification problems such as binary and multi-class classification, each input sample corresponds to a single target label. Therefore the classifier is required to identify the single target label corresponding to the input sample. On the contrary, in multi-label classification, each of the input samples belongs to a subset of target labels. Therefore, the multi-label classifier is required to identify both the number of labels and the identity of the labels in order to perform multi-label classification. This results in the increased complexity of the multi-label classification problems. Another key challenge in implementing a generic multi-label classifier is that, not all datasets are equally multi-labelled. The degree of multi-labelness varies for every dataset. The increased complexity and the varying degree of multi-labelness are the two major challenges in developing a multi-label classifier.

The proposed method falls under the algorithm adaptation techniques category as the base algorithm is extended to adapt to the multi-label problem. The various steps involved in the proposed online sequential multi-label ELM (OSML-ELM) approach are

- Initialization
- Pre-processing
- ELM Training
- ELM Testing
- Multi-label identification

Pre-processing and post-processing of data are of prime importance in extending the ELM based technique for multi-label classification.

Initialization: The fundamental parameters of the ELM network such as the number of hidden layer neurons and the activation function are initialized. The number of hidden layer neurons is selected for each dataset so as to avoid the over-fitting problem. The input weights and the bias value of the network are randomly initialized.

Pre-processing: In single label classification, each of the input samples corresponds to only one target class. Therefore, the dimension of the target output label is always fixed at 1. On the contrary, in multi-label classification, each input is associated with an M-tuple output label with each element of the set as 0 or 1 representing the belongingness of the input corresponding to the target labels. Therefore the dimension of the target output label is 'M'. The label set denoting the belongingness for each of the labels is converted from unipolar representation to bipolar representation.

ELM Training: The processed input is then supplied sequentially to the online sequential variant of the ELM technique. Let $N_0$ be the number of samples in the initial block, from equations (9) and (10), the initial output weight $\beta_0$ is calculated as $\beta_0 = M_0 H_0^T Y$ where $M_0 = (H_0^T H_0)^{-1}$.

Upon calculating the initial output weight $\beta_0$, for each sequentially arriving data/block of data, the output weights of the network are updated based on recursive least square using the equations,

$$M_{k+1} = M_k - \frac{M_k h_{k+1} h_{k+1}^T M_k}{1 + h_{k+1}^T M_k h_{k+1}} \quad (11)$$

$$\beta_{k+1} = \beta_k + M_{k+1} h_{k+1}(Y_{k+1}^T - h_{k+1}^T \beta_k) \quad (12)$$

where $k = 0, 1, 2, \ldots N-N_0-1$.

The theory and the mathematics behind using the recursive least square for online sequential ELM is discussed in several papers [41, 42] in the literature.

---

**Algorithm: Proposed OSML-ELM algorithm for multi-label classification**

1. Initialization: The fundamental parameters of the network are initialized
2. Pre-processing: The raw input data is processed for classification
3. ELM Training:
   – Initial phase
     Processing of initial block of data
     $M_0 = (H_0^T H_0)^{-1}$    $\beta_0 = M_0 H_0^T Y_0$
   – Sequential phase
     Online processing of sequential data
     $M_{k+1} = M_k - \frac{M_k h_{k+1} h_{k+1}^T M_k}{1 + h_{k+1}^T M_k h_{k+1}}$
     $\beta_{k+1} = \beta_k + M_{k+1} h_{k+1}(Y_{k+1}^T - h_{k+1}^T \beta_k)$
   – Threshold identification
4. ELM Testing:
   Estimation of raw output values using $Y = H\beta$
5. Post-processing and multi-label identification
   The raw output values is compared with the threshold value.
   Separation into two categories of labels (Labels that the data sample belong to and labels the data sample does not belong to)
   Identifying the number of labels corresponding to input data sample
   Identifying the target class labels for the input data sample

---

ELM Testing: During the testing phase, the network computes the predicted raw output value Y using the formula $Y = H\beta$ where $\beta$ is output weight obtained during the training phase.

Multi-label Identification: The multi-label identification step is the key step in extending the ELM based technique for multi-label classification. As foreshadowed, multi-label classifiers are required to predict both the number of target labels and the identity of the target labels corresponding to each of the input samples. Since the number of labels corresponding to each input is completely unknown and dynamic, a thresholding based technique is used. The threshold value is selected during the training phase such that it maximizes the separation between the family of labels the input belongs to and the family of labels the input does not belong to. Setting up of the threshold value is of prime importance as it directly affects the performance of the classifier. The raw output values Y obtained from the previous step is then compared to the threshold value. The number of raw output values that are greater than the threshold determines the number of target labels corresponding the input sample and the index of the corresponding values determines the identity of the target labels. The overview of the proposed algorithm is summarized.

## IV. EXPERIMENTATION

This section elaborates the experimental design and the dataset specifications used to evaluate the performance of the proposed technique. Multi-label problems have a unique feature called the degree of multi-labelness. In other words, not all datasets are equally multi-labelled. The multi-label nature of datasets varies widely from each other. The differences in the number of labels, the number of samples having multiple labels and the average number of target labels for each sample result in the varying degree of multi-labelness to each dataset. Two metrics, label cardinality (LC) and label density (LD) are used to quantitatively measure the degree of multi-labelness of the datasets. Label cardinality is the average number of labels corresponding to each sample in the dataset. Label density also factors the number of labels in the dataset in addition to the average number of labels [23]. The LC and LD can be calculated using the following equations.

$$Label - Cardinality = \frac{1}{N} \sum_{i=1}^{N} |Y_i| \quad (13)$$

$$Label - Density = \frac{1}{N} \sum_{i=1}^{N} \frac{|Y_i|}{|L|} \quad (14)$$

where, N is the number of training samples, L is the label set and Yi gives the multi-label belongingness to target labels corresponding to each input sample. The impact of differences in LC and LD and its influence on the performance of the classifier is discussed by Bernardini et. al in [43]. Two datasets have same LC and different LD or vice versa can result in significant variation in the performance of the classifier [16].

The proposed technique is experimented with five datasets from different application domains and wide range of LC and LD. The specifications of the dataset used are tabulated in Table 1. The performance metrics such as hamming loss, accuracy, F1 measure, training time and testing time are evaluated for the five datasets and the results are compared with five state-of-the-art techniques. The details of the state-of-the-art techniques are given in Table 2.

TABLE 1: DATASET SPECIFICATIONS

| Dataset | Domain | No. of Labels | No. of Features | No. of Samples | LC | LD |
|---|---|---|---|---|---|---|
| Yeast | Biology | 14 | 103 | 2417 | 4.24 | 0.303 |
| Scene | Multi-media | 6 | 294 | 2407 | 1.07 | 0.178 |
| Corel5k | Multi-media | 374 | 499 | 5000 | 3.52 | 0.009 |
| Enron | Text | 53 | 1001 | 1702 | 3.38 | 0.064 |
| Medical | Text | 45 | 1449 | 978 | 1.25 | 0.027 |

TABLE 2: METHODS USED FOR COMPARISON

| Method Name | Machine Learning Category | Method Category |
|---|---|---|
| QWeighted approach for Multi-label Learning (QWML) | Support Vector Machine | Problem Transformation |
| Predictive Clustering Trees (PCT) | Decision Trees | Algorithm Adaptation |
| Multi-Label k-Nearest Neighbors (ML-kNN) | Nearest Neighbors | Algorithm Adaptation |
| Random Forest Predictive Clustering Trees (RF-PCT) | Decision Trees | Ensemble |
| Random Forest of ML-C4.5 (RFML-C4.5) | Decision Trees | Ensemble |

## V. RESULTS AND DISCUSSIONS

The proposed technique is evaluated in terms of consistency, performance and compatibility for large scale streaming data applications.

### A. Performance Metrics

The complex nature of multi-label problems results in a unique feature of multi-label classifiers called partial correctness of results. Since both the number of labels and the label identities are to be predicted by the multi-label classifier, the problem of partial correctness arises. The classifier can wrongly predict either the number of labels corresponding to an input sample is the identity of corresponding target labels. Therefore, the hamming loss performance metric is used to quantitatively evaluate the performance of the classifier along with accuracy, precision, recall and F1 measure. The hamming loss is calculated as the fraction of wrong labels to the total number of labels. Hamming loss for an ideal classifier is zero. Also, the training time and the testing time of the proposed technique is evaluated and compared with the state-of-the-art techniques.

The results for the performance metrics by the proposed technique is tabulated in Table 3. The proposed technique is compared with five different state-of-the-art techniques and the results are given in Tables 4-5 and Fig. 2-3.

TABLE 3: PERFORMANCE OF OSML-ELM

| Dataset | Hamming Loss | Accuracy | Precision | Recall | F1 measure | Training time | Testing time |
|---|---|---|---|---|---|---|---|
| Yeast | 0.206 | 0.493 | 0.693 | 0.580 | 0.632 | 0.114 | **0.017** |
| Scene | 0.098 | 0.610 | 0.630 | 0.645 | 0.637 | 2.329 | **0.047** |
| Corel5k | 0.009 | 0.060 | 0.175 | 0.063 | 0.093 | 5.365 | **0.076** |
| Enron | 0.049 | 0.404 | 0.640 | 0.461 | 0.536 | 0.630 | **0.028** |
| Medical | 0.011 | 0.713 | 0.760 | 0.740 | 0.750 | 0.663 | **0.039** |

TABLE 4: COMPARISON OF TRAINING TIME (S)

| Dataset | QWML | PCT | ML-kNN | RFML-C4.5 | RF-PCT | OSML-ELM |
|---|---|---|---|---|---|---|
| Yeast | 672 | 1.5 | 8.2 | 19 | 25 | **0.114** |
| Scene | 195 | 2 | 14 | 10 | 23 | **2.329** |
| Corel5k | 2388 | 30 | 389 | 385 | 902 | **5.365** |
| Enron | 971 | 1.1 | 6 | 25 | 47 | **0.630** |
| Medical | 40 | 0.6 | 1 | 7 | 27 | **0.663** |

TABLE 5: COMPARISON OF TESTING TIME (S)

| Dataset | QWML | PCT | ML-kNN | RFML-C4.5 | RF-PCT | OSML-ELM |
|---|---|---|---|---|---|---|
| Yeast | 64 | 0 | 5 | 0.5 | 0.2 | **0.017** |
| Scene | 40 | 0 | 14 | 2 | 1 | **0.047** |
| Corel5k | 119 | 1 | 45 | 1.8 | 2.5 | **0.076** |
| Enron | 174 | 0 | 3 | 1 | 1 | **0.028** |
| Medical | 25 | 0 | 0.2 | 0.5 | 0.5 | **0.039** |

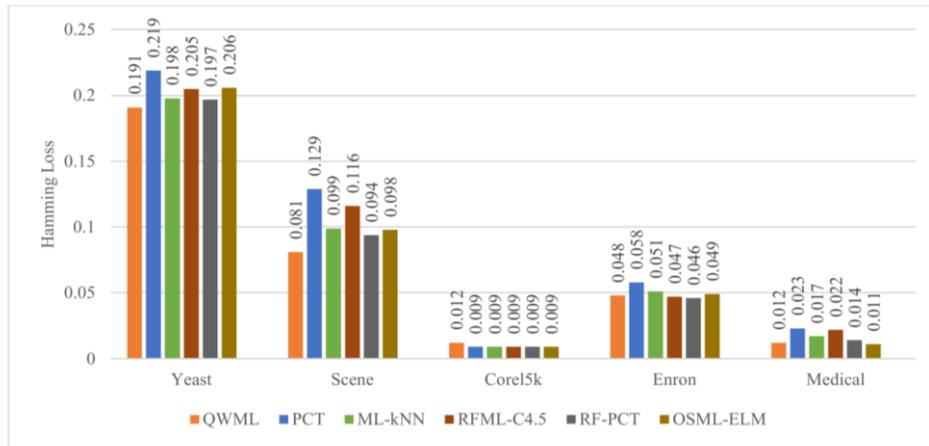

Fig. 2. Comparison of Hamming Loss

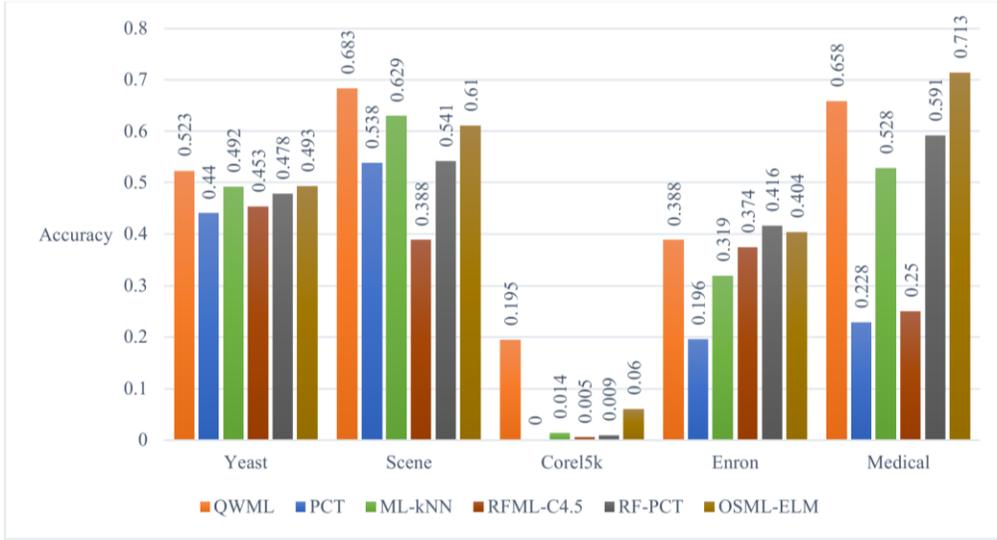

Fig. 3. Comparison of Accuracy

From the results it can be clearly seen that the proposed technique outperforms all the existing techniques in terms of speed and performance.

*B. Consistency*

Consistency is the key virtue that is essential for any new technique. Since the initial weights of the networks are randomly chosen for an ELM based technique, evaluation of consistency is of critical importance. Therefore the proposed technique is evaluated for consistency using 5-fold and 10-fold cross validation. The results obtained are tabulated in Table 6.

TABLE 6: CONSISTENCY

| Dataset | Hamming Loss – 5-fcv | Hamming Loss – 10-fcv |
|---|---|---|
| Yeast | 0.206 ± 0.001 | 0.206 ± 0.002 |
| Scene | 0.098 ± 0.002 | 0.098 ± 0.002 |
| Corel5k | 0.009 ± 0.000 | 0.009 ± 0.000 |
| Enron | 0.049 ± 0.001 | 0.049 ± 0.001 |
| Medical | 0.011 ± 0.001 | 0.011 ± 0.001 |

*C. Streaming Data Classification*

In order to perform real-time streaming data classification, the time taken to process one chunk of sequentially arriving streaming data is of critical importance. If the time taken for executing a single block of data stream is more than the arrival rate of the data stream, real-time processing cannot be achieved. The time taken by the proposed technique for a single block of data stream can be calculated using the training time and the number of blocks during the training phase. The results are summarized in Table 7.

TABLE 7: AVERAGE TIME PER BLOCK OF DATA

| Dataset | Training Time (s) | Number of blocks | Average time(s)/block |
|---|---|---|---|
| Yeast | 0.114 | 51 | 0.00223529 |
| Scene | 2.329 | 48 | 0.04852083 |
| Corel5k | 5.365 | 93 | 0.05768817 |
| Enron | 0.63 | 48 | 0.013125 |
| Medical | 0.663 | 37 | 0.01791892 |

From the table it is evident that the proposed technique is capable of performing multi-label classification in real-time for streaming data.

## VI. CONCLUSION

The proposed OSML-ELM technique is a real-time online multi-label classifier for streaming data applications. The performance of the proposed method is experimented on five datasets of different domains with a wide range of LC and LD. From the results it is evident that the proposed method is consistent and outperforms the existing state-of-the-art techniques in terms of speed and remains one of the top methods in terms of performance. The high-speed nature of the OSML-ELM supports scalability of the proposed technique for real-time data streams.

ACKNOWLEDGEMENT

The authors would like to acknowledge the funding support from the Ministry of Education, Singapore (Tier 1 AcRF, RG30/14), Rajasekar Venkatesan is supported by NTU Research Student Scholarship.